\documentclass[paper]{ieice}
\usepackage{graphicx,xcolor}
\usepackage[fleqn]{amsmath}
\usepackage{newtxtext}
\usepackage[varg]{newtxmath}
\usepackage{mathtools}
\usepackage{algorithmic}
\usepackage{algorithm}
\usepackage[hang,small,bf]{caption}
\usepackage[subrefformat=parens]{subcaption}
\field{}
\title{Extrinsicaly Rewarded Soft Q Imitation Learning with Discriminator}
\authorlist{%
 \authorentry[rfuruyama@csl.ec.t.kanazawa-u.ac.jp]{Ryoma FURUYAMA}{Kanazawa University}{}\MembershipNumber{1}
\authorentry[dkuyoshi@csl.ec.t.kanazawa-u.ac.jp]{Daiki KUYOSHI}{Kanazawa University}{}\MembershipNumber{2}
\authorentry[syamane@is.t.kanazawa-u.ac.jp]{Satoshi YAMANE}{Kanazawa University}{}\MembershipNumber{3}
}
\affiliate[affiliate label]{The author is with the 
}

\received{2015}{1}{1}
\revised{2015}{1}{1}

\begin{document}
\maketitle
\begin{summary}
Imitation learning is often used in addition to reinforcement learning in environments where reward design is difficult or where the reward is sparse, but it is difficult to be able to imitate well in unknown states from a small amount of expert data and sampling data. Supervised learning methods such as Behavioral Cloning do not require sampling data, but usually suffer from distribution shift. The methods based on reinforcement learning, such as inverse reinforcement learning and Generative Adversarial imitation learning (GAIL), can learn from only a few expert data. However, they often need to interact with the environment. Soft Q imitation learning (SQIL) addressed the problems, and it was shown that it could learn efficiently by combining Behavioral Cloning and soft Q-learning with constant rewards. In order to make this algorithm more robust to distribution shift, we propose more efficient and robust algorithm by adding to this method a reward function based on adversarial inverse reinforcement learning that rewards the agent for performing actions in status similar to the demo. We call this algorithm Discriminator Soft Q Imitation Learning (DSQIL). We evaluated it on MuJoCo environments.
\end{summary}

\maketitle

\section{Introduction}
  Recent developments in the field of deep reinforcement learning have made it possible to learn diverse behaviors for high-dimensional input. However, there are still some problems. Among them, we focus on the efficiency of learning and the difficulty of designing a reward function \cite{boyd2004convex}. For example, when using reinforcement learning for artificial intelligence of autonomous driving, it is necessary to deal with many unexpected phenomena such as various terrain and people coming out \cite{sumanth2021enhanced}. If we design a reward function to solve this problem, the program may become enormous. Also, incomplete reward function design may promote unexpected behavior. In addition, it is necessary to explore with many random actions until the agent obtains a reward in an environment with sparse rewards.
  
  In such setting of problems, imitation learning is often used instead of reinforcement learning. Behavioral Cloning \cite{pomerleau1991efficient}, which is the classical imitation learning, is a simple supervised learning algorithm that maximizes the likelihood of the actions taken by an expert in a certain state. It shows good results for simple tasks, but it requires a large dataset of the pairs of state and action, and it sometimes behaves strangely in a state is not in the dataset. In order to overcome these disadvantages, Inverse Reinforcement Learning (IRL) \cite{ng2000algorithms} performs a two-step learning process in which it estimates a reward function instead of expert actions, and it performs reinforcement learning based on the reward function. This algorithm helps to learn to behave in unexpected situations.

  However, IRL has the disadvantage of being unstable due to two stages of learning. Therefore, the methods of learning the behavior of the expert directly by Generative Adversarial Networks (GANs) \cite{goodfellow2014generative} without explicitly finding a reward function were proposed such as Generative Adversarial imitation Learning (GAIL) \cite{ho2016generative}. This methods can learn efficiently with small amounts of data. Furthermore, Adversarial Inverse Reinforcement Leaning (AIRL) \cite{fu2017learning} has been proposed for IRL, which outperforms IRL and GAIL by improving generalizability through the use of GAN. Thus, imitation learning has been greatly advanced by GANs, and methods and applications based on this technology continue to be studied \cite{choi2021trajgail,wu2021textgail}.
  
  Although Behavioral Cloning was no longer considered useful, Reddy et al. \cite{reddy2019sqil} proposed Soft Q Imitation Learning (SQIL), which addressed state distribution shift by the combination of Behavioral Cloning and reinforcement learning. It has been reported that the learning has been performed efficiently with less training steps than in previous adversarial imitation learning. In this paper, we propose more efficient and robust algorithm by adding to this method a reward function that rewards the agent for performing actions in status similar to the demo. We evaluate it with three environments of MuJoCo \cite{todorov2012mujoco}, and we show the strong and weak points.

\section{Background}
We consider problems that satisfy the definition of Markov Decision Process (MDP) \cite{sutton2018reinforcement}. In continuing tasks, the returns for a trajectory $\tau=\left(s_{t},a_{t}\right)_{t=0}^\infty$ are defined as $r_{t}=\sum_{k=t}^{\infty}\gamma^{k-t}R\left(s_{k},a_{k}\right)$ , where $\gamma$ is a discount factor. In order to use the same notation for episodic tasks, we can define a set of absorbing state $s_{a}$. When we define the reward $R\left(s_{a}, \cdot\right)=0$, we can define returns simply as $r_{t}=\sum_{k=t}^{T}\gamma^{k-t}R\left(s_{k},a_{k}\right)$. In reinforcement learning like Actor Critic (AC) \cite{konda1999actor} or Q-learning \cite{watkins1992q}, we would like to learn a policy $\pi$ that maximizes expected returns.Therefore, the objective function is
\begin{equation}
  \label{RL_obj}
  J\left(\pi\right)=\sum_{t=0}^T\mathbb{E}_{\left(s_t,a_t\right)\sim\rho_{\pi}}\left[r\left(s_t,a_t\right)\right].
\end{equation}

Recently, various methods have been studied, including AC-based Proximal Policy Optimization (PPO) \cite{schulman2017proximal} and Q-learning-based Recurrent Replay Distributed DQN (R2D2) \cite{kapturowski2018recurrent}. One of them is Soft Actor Critic (SAC) \cite{haarnoja2018soft,haarnoja2018softa} proposed by Haarnoja et al. SAC is a maximum entropy reinforcement learning and has shown excellent performance, especially in complex environments. Maximum entropy reinforcement learning significantly improves explorability and robustness. is one of the methods. The objective function is Equation \ref{RL_obj} plus an entropy maximization term $H\left(\pi\left(\cdot|s_t\right)\right)$, 

\begin{equation}
  \label{SAC_obj}
  J\left(\pi\right)=\sum_{t=0}^T\mathbb{E}_{\left(s_t,a_t\right)\sim\rho_{\pi}}\left[r\left(s_t,a_t\right)+\alpha H\left(\pi\left(\cdot|s_t\right)\right)\right],
\end{equation}
where $\alpha$ is the temperature parameter, which determines the relative importance of the entropy term to the reward and controls the stochasticity of the optimal policy. SAC uses a soft Q function \cite{haarnoja2017reinforcement} to learn measures that maximize estimates while improving their accuracy. This method optimizes stochastic policies off-policy, and this algorithm is sample efficient.

\section{Related works}
\subsection{Behavioral Cloning}
Behavioral Cloning \cite{pomerleau1991efficient} is a classical imitation learning algorithm. It is a method of supervised learning that takes the state $s_E$ in the expert data as input and regards the action $a_E$  of the expert as the label.When the set of expert state and action pairs $\left(s_E,a_E\right)$ is $\beta_{demo}$, the loss function for the parameter $\theta$ is the following equation.
\begin{equation}
\label{BC_loss}
l_{BC}\left(\theta\right)=\sum_{\left(s,a\right)\in\beta_{demo}}-\log{\pi_{\theta}\left(a\mid s\right)},
\end{equation}
where $\pi_{\theta}$ is a measure with parameter $\theta$, and the objective is to find $\theta$ that minimizes Equation \ref{BC_loss}. Although this method is simple, it is easy to overfit to the expert data because it does not learn the result of the action, and it suffers from the state distribution shift. As a result, it has the disadvantage of not being able to make good decisions for unseen states \cite{ross2011reduction}.

\subsection{Soft Q Imitation Learning}
To cover the shortcomings of behavioral cloning, Reddy et al \cite{reddy2019sqil} proposed Soft Q Imitation Learning (SQIL). SQIL combines behavioral cloning and reinforcement learning to address shifts in the state distribution. This method is built on soft Q learning \cite{haarnoja2017reinforcement}, where experts are assumed to follow a policy $\pi$ that maximizes reward $R\left(s,a\right)$ in an infinite horizon Markov decision process (MDP) with continuous state space S and discrete action space A. The policy $\pi\left(a\mid s\right)$ forms a Boltzmann distribution for action,
\begin{equation}
\label{policy}
\pi\left(a\mid s \right)\triangleq\frac{\exp\left(Q\left(s,a\right)\right)}{\sum_{a'\in A}\exp\left(Q\left(s,a'\right)\right)},
\end{equation}
where $Q$ is soft Q function and is defined by the following equation.

\begin{equation}
\label{Q_function}
Q\left(s,a\right)\triangleq R\left(s,a\right)+\gamma\mathbb{E}_{s'}\left[\log\left({\sum_{a'\in A}\exp{Q\left(s',a'\right)}}\right)\right]
\end{equation}
$s'$ is the state when action $a$ is taken in state $s$. If we assume that the behavior of our agent follows the policy, we can define the loss function by Equation \ref{BC_loss}.

\begin{equation}
\begin{split}
\label{BC_loss_1}
l_{BC}\left(\theta\right)\triangleq \sum_{\left(s,a\right)\in\beta_{demo}}&-\left(Q_{\theta}\left(s,a\right)\right. \\
&\left.-\log{\left(\sum_{a'\in A}\exp\left(Q_\theta\left(s,a\right)\right)\right)}\right)
\end{split}
\end{equation}
SQIL aims to learn by considering the trajectory by regularizing this loss function with squared soft Bellman error $\delta^2\left(\beta,r\right)$, called Regularized Behavioral Cloning (RBC) \cite{piot2014boosted}.

\begin{equation}
\label{RBC_loss}
l_{RBC}\left(\theta\right)\triangleq l_{BC}\left(\theta\right)+\lambda\delta^2\left(\beta_{demo}\cup\beta_{samp}, 0\right)
\end{equation}

\begin{equation}
\begin{split}
\label{soft_bellman_error}
\delta^2\left(\beta, r\right)\triangleq \frac{1}{\mid\beta\mid}
  & \sum_{\left(s,a,s'\right)\in\beta}\left(Q_\theta\left(s,a\right) \right.\\
  & \quad \left.-\left(r+\gamma \log \left(\sum_{a'\in A}\exp\left(Q_\theta\left(s',a'\right)\right)\right)\right)\right)^2 , 
\end{split}
\end{equation}
where $\lambda \in R_{\geq 0}$ is a hyperparameter that determines the relative importance of Behavioral Cloning versus soft Q learning. In addition, $\beta_{demo}$ and $\beta_{samp}$ are a replay buffer of demonstration data by an expert and sampling data from an environment by an agent respectively.

Furthermore, we can rewrite the gradient of $l_{RBC}\left(\theta\right)$ as simple form.
\begin{equation}
\begin{split}
\label{grad_RBC}
\bigtriangledown_\theta l_{RBC}\left(\theta\right)\propto\bigtriangledown_\theta\left( \right.
  & \delta^2\left(\beta_{demo}, 1\right)+\lambda_{samp}\delta^2\left(\beta_{samp},0\right)\\
  & \left. +\log\left(\sum_{a\in A}\exp\left(Q_\theta\left(s_0,a\right)\right)\right)\right) , 
\end{split}
\end{equation}
where $s_0$ is the initial state. The update equation for $\theta$ in the SQIL algorithm is defined by Equation \ref{grad_RBC}.

\begin{equation}
\label{sqil_update}
\theta\leftarrow\theta - \eta\triangledown_{\theta}\left(\delta^2\left(\beta_{demo}, 1\right)+\lambda_{samp}\left(\beta_{samp}, 0\right)\right)
\end{equation}
In original paper, $\lambda_{samp}=1$. Importantly, we can recognize that the SQIL agent sets the rewards of all demonstration data to 1 and the rewards of all sampling data to 0.

SQIL can be implemented because it requires only minor modifications to standard Q-learning implementations; SQIL can also be extended to MDPs with continuous action spaces by simply replacing Q-learning with an off-policy actor-critic method such as SAC Given the difficulty of correctly implementing deep RL algorithms \cite{henderson2018deep}, this flexibility is an advantage and enhances the utility of SQIL because it can be built on top of existing implementations of deep RL algorithms.

\subsection{Generative Adversarial Networks}
\begin{figure}[h]
  \centering
  \includegraphics[keepaspectratio, scale=0.35]{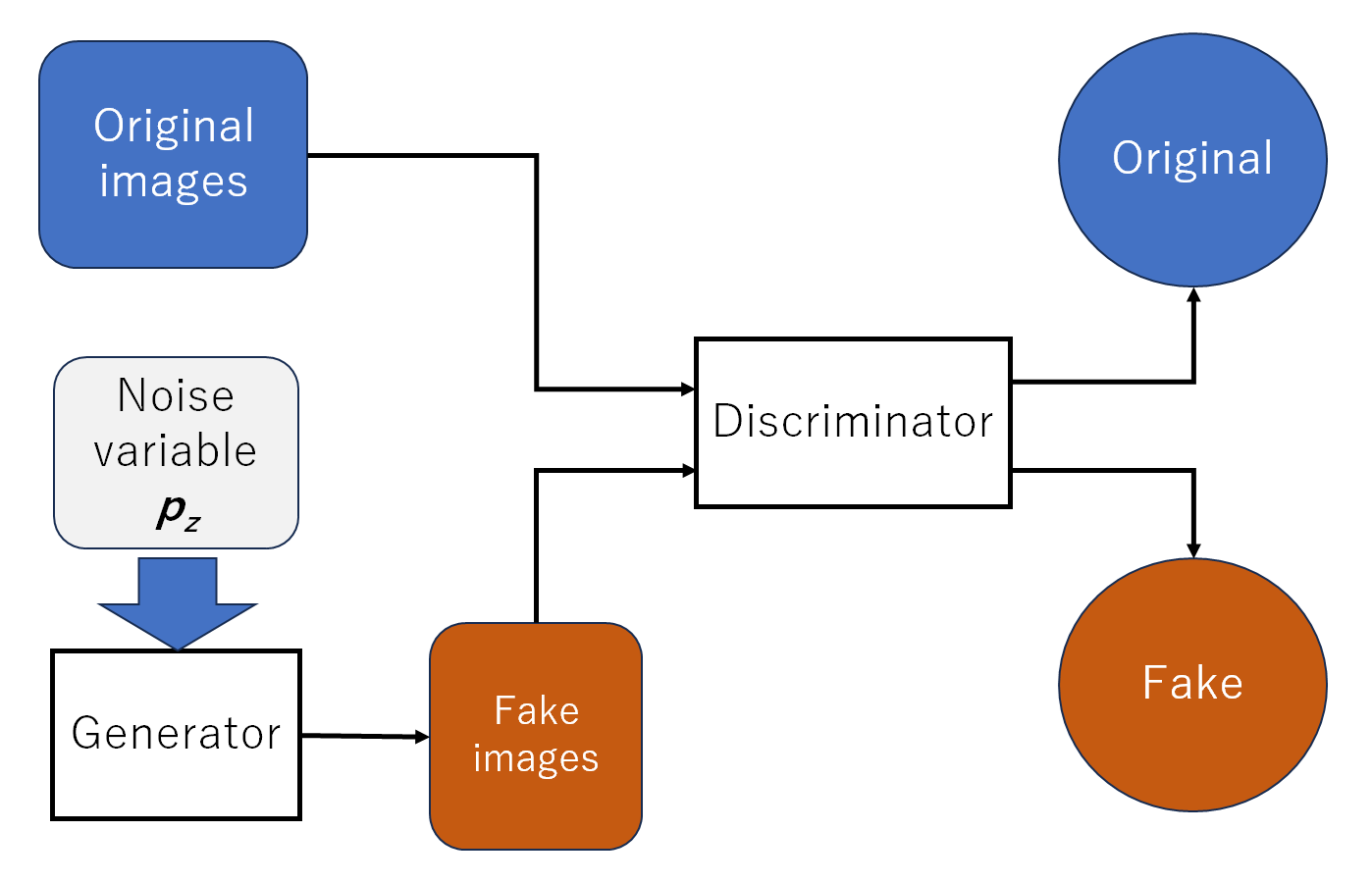}
  \caption{Generative Adversarial Network Architecture.}
  \label{figGANs}
\end{figure}

Generative Adversarial Networks (GANs) \cite{goodfellow2014generative} is one of the methods used for image generation. In this method, two models are prepared: a generator $G$ and a discriminator $D$. In the generator $G$, the distribution of the input noise variable $p_z$ is predefined and the prepared data is trained. The objective of this learning is to generate data similar to the training example from the noise variables. The objective of $D$, on the other hand, is to discriminate between the data in the training example and the data generated in $G$. In this way, $G$ and $D$ learn to compete with each other. In other words, $G$ and $D$ are playing a mini-max game based on the value function $V\left(G,D\right)$.
\begin{equation}
\begin{split}
\label{GANs}
\min_G\max_DV\left(G,D\right)=
  & \mathbb{E}_{x\sim p_{data}\left(x\right)}\left[\log D\left(x\right)\right] \\
  & +\mathbb{E}_{z\sim p_z\left(z\right)}\left[\log\left(1-D\left(G\left(z\right)\right)\right)\right],
\end{split}
\end{equation}
where $p_{data}$ is the distribution of the data in the train example.Although GANs are image generators, they are well suited for reinforcement learning and have contributed greatly to the advancement of that technology, especially in imitation learning; GAIL \cite{ho2016generative} is one such example. An overview of the architecture of GANS is shown in Figure \ref{figGANs}.

\section{Proposal of Discriminator Soft Q Imitation Learning}

\subsection{Discriminator Soft Q Imitation Learning}
\begin{figure*}[t]
  \centering
  \includegraphics[keepaspectratio, scale=0.55]{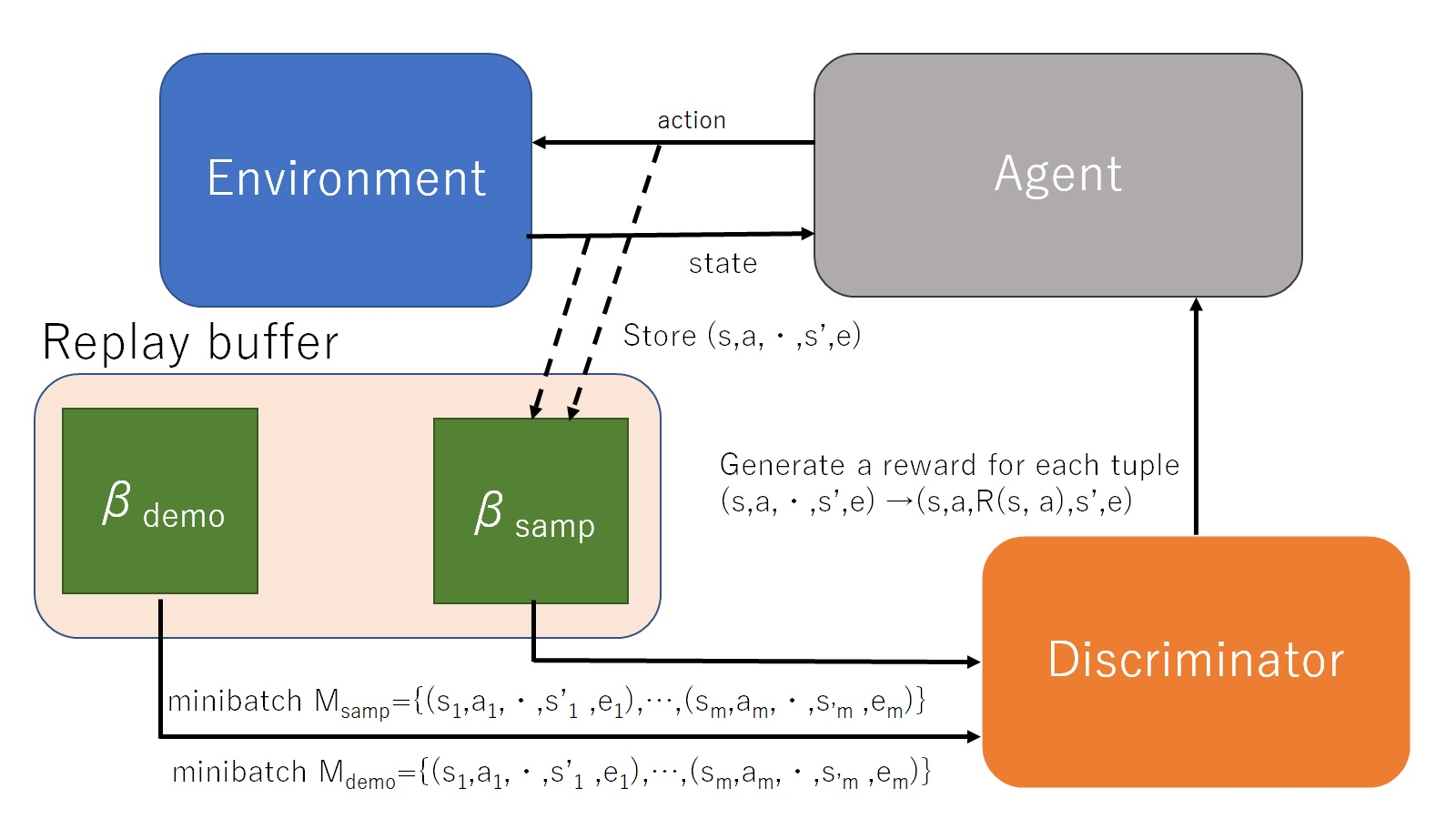}
  \caption{The overall of DSQIL algorithm.}
  \label{figDSQIL}
\end{figure*}

SQIL \cite{reddy2019sqil} shows that soft Q learning, which provides a positive constant reward for expert data, can efficiently mimic an agent. However, the method of determining this constant reward is not always a good method. For example, after some progress in learning, the agent learns as reward 0 even if the sample data obtained from the environment is similar to the expert's data. This may become noise in the learning process. Therefore, we propose a method that uses the discriminator D of the GAN as the reward function. This method is expected to reduce the above problem as well as to learn efficiently with less expert data. We call this method Discriminator Soft Q Imitation Learning (DSQIL). DSQIL's algorithm is shown in Algorithm \ref{DSQIL}, and an overall view of DSQIL is shown in Figure \ref{figDSQIL}. As with SQIL, the agent can be used flexibly, for example, Q-Learning \cite{watkins1992q} for discrete value control tasks and SAC \cite{haarnoja2018soft} for continuous value control tasks.

\begin{figure}[h]
\begin{algorithm}[H]
    \caption{Discriminator Soft Q Imitation Learning (DSQIL)}
    \label{DSQIL}
    \begin{algorithmic}[1]
    \REQUIRE Replay buffer of demonstaration data $\beta_{demo}$,\\
    $ \lambda_{demo} = 1, \lambda_{samp} = 1$ in our experiment.
    \STATE Initialize replay buffer of sample data $\beta_{samp}\leftarrow\emptyset$
    \FOR{$n = 1,2,...$}
    \WHILE{$e \neq True$}
    \STATE $\tau = \left(s,a,.,s',e\right)$ with $\pi_\theta$
    \STATE $\beta_{samp} \leftarrow \beta_{samp} \cup \tau$
    \STATE $M_{demo} = \left\{\left(s_t,a_t,.,s'_t,e_t\right)\right\}_{t=1}^{m}\sim\beta_{demo}$
    \STATE $M_{samp} = \left\{\left(s_t,a_t,.,s'_t,e_t\right)\right\}_{t=1}^{m}\sim\beta_{samp}$
    \STATE Calcurate the loss of $D$
    \STATE Update $D$ with GAN
    
    \FOR{$i = 1,2,...,M_{demo}$}
    \STATE $R\left(\beta_{demo_i}\right) \leftarrow \frac{D\left(s_i,a_i\right)}{2}+\frac{1}{2\lambda_{demo}}$
    \ENDFOR
    \FOR{$j = 1,2,...,M_{samp}$}
    \STATE $R\left(\beta_{samp_j}\right) \leftarrow \frac{D\left(s_j,a_j\right)}{2}$
    \ENDFOR
    \STATE Update $\theta$
    \COMMENT{See Equation \ref{DSQIL_update}}
    \ENDWHILE
    \ENDFOR
    
    \end{algorithmic}
\end{algorithm}
\end{figure}

\subsection{Update equation}
We set up the DSQIL update equation based on Equation \ref{sqil_update}.

\begin{equation}
\begin{split}
\label{DSQIL_update}
\theta\leftarrow\theta - \eta\triangledown_{\theta}\left(\right.
  &\lambda_{demo}\delta^2\left(\beta_{demo}, R\left(\beta_{demo}\right)+\frac{1}{2\lambda_{demo}}\right)\\
  &\left.+\lambda_{samp}\delta^2\left(\beta_{samp}, R\left(\beta_{samp}\right)\right)\right),
\end{split}
\end{equation}
where $\lambda_{demo} \in \mathbb{R}_{\geq0}, \lambda_{samp} \in \mathbb{R}_{\geq0}$ are hyperparameter, and $\delta^2$ denotes the squared soft Bellman error defined in Equation \ref{soft_bellman_error}. We added to the rewards, in addition to the fixed values, a function that indicates the probability of similarity of the expert data using discriminator $D$ from GANs \cite{goodfellow2014generative}. This change enabled more efficient learning by rewarding sample data similar to the expert data.

The update equation was derived from the following loss function defined with reference to Equation \ref{RBC_loss}.

\begin{equation}
\label{loss_DSQIL}
l_{DSQIL}\left(\theta\right)\triangleq l_{BC}\left(\theta\right)+\lambda\delta^2\left(\beta_{demo}\cup\beta_{samp},R\right),
\end{equation}
where, $R$ is reward function.In addition, the soft value function is defined as follows.

\begin{equation}
\label{soft_value_function}
V\left(s\right)\triangleq \log\left(\sum_{a\in A}\exp\left(Q_{\theta}\left(s,a\right)\right)\right)
\end{equation}

In Equation \ref{loss_DSQIL}, dividing the soft Bellman squared error term by demonstration data and sample data, and further using Equation \ref{BC_loss_1},

\begin{equation}
\begin{split}
\label{grad_dsqil}
\bigtriangledown&
  l_{DSQIL}\left(\theta\right)=\sum_{\tau\in\beta_{demo}}\sum_{t=0}^{T-1}-\left(\bigtriangledown Q_{\theta}\left(s_t,a_t\right)-\bigtriangledown V\left(s_t\right)\right)\\
  &+\lambda_{demo}\sum_{\tau\in\beta_{demo}}\sum_{t=0}^{T-1}\bigtriangledown\left(Q_\theta\left(s_t,a_t\right)-\left(R+\gamma V\left(s_{t+1}\right)\right)\right)^2\\
  &+\lambda_{samp}\bigtriangledown\delta^2\left(\beta_{samp},R\right)\\
  &=\sum_{\tau\in\beta_{demo}}\sum_{t=0}^{T-1}\left(V\left(s_t\right)-\gamma V\left(s_{t+1}\right)\right)\\
  &+\lambda_{demo}\bigtriangledown\delta^2\left(\beta_{demo},R+\frac{1}{2\lambda_{demo}}\right)\\
  &+\lambda_{samp}\bigtriangledown\delta^2\left(\beta_{samp},R\right).
\end{split}
\end{equation}
Assuming $\gamma\triangleq1$, the inner product of the first term is a telescoping sum.

\begin{equation}
\begin{split}
\label{grad_dsqil_1}
\bigtriangledown l_{DSQIL}\left(\theta\right)&=\sum_{\tau\in\beta_{demo}}\left(\bigtriangledown V\left(s_0\right)-\triangledown V\left(s_T\right)\right)\\
  &+\lambda_{demo}\bigtriangledown\delta^2\left(\beta_{demo},R+\frac{1}{2\lambda_{demo}}\right)\\
  &+\lambda_{samp}\bigtriangledown\delta^2\left(\beta_{samp},R\right)
\end{split}
\end{equation}
From the assumption that $s_T$ is absorbed, $V\left(s_T\right)=0$. Therefore,

\begin{equation}
\begin{split}
\label{grad_dsqil_2}
\bigtriangledown l_{DSQIL}\left(\theta\right)&=\sum_{\tau\in\beta_{demo}}\bigtriangledown V\left(s_0\right)\\
  &+\lambda_{demo}\bigtriangledown\delta^2\left(\beta_{demo},R+\frac{1}{2\lambda_{demo}}\right)\\
  &+\lambda_{samp}\bigtriangledown\delta^2\left(\beta_{samp},R\right).
\end{split}
\end{equation}
In our experiments, all demo rollouts start from the same initial state s0. Thus,
\begin{equation}
\begin{split}
\label{grad_dsqil_3}
\bigtriangledown l_{DSQIL}\left(\theta\right)&\propto\bigtriangledown\left(\lambda_{demo}\delta^2\left(\beta_{demo},R+\frac{1}{2\lambda_{demo}}\right)\right.\\
  &\left.+\lambda_{samp}\delta^2\left(\beta_{samp},R\right)+V\left(s_0\right)\right).
\end{split}
\end{equation}
Thus, the gradient of the loss function is similar to the updated equation shown in Equation \ref{DSQIL_update} plus the soft value at $s_0$.

\subsection{Reward function}
In the update equation, we had placed the reward function $R$. We used the discriminator $D$ with reference to GANs \cite{goodfellow2014generative} as the reward function R.The discriminator $D$ was used as the reward function. This discriminator $D$ is trained to discriminate between expert data and sample data, and the probability of being expert data is expressed as a value between 0 and 1. In other words, the value is close to 1 if it discriminates the data as expert data and close to 0 if it judges the data as other than expert data. To optimize the discriminator $D$ with weight $\phi$, We minimize the following loss function,
\begin{equation}
\begin{split}
\label{Reward_DSQIL}
  l\left(\phi\right)\triangleq-&\mathbb{E}_{\left(s,a\right)\sim\beta_{demo}}\left[\log\left(D_\phi\left(s,a\right)\right)\right]\\
  &-\mathbb{E}_{\left(s,a\right)\sim\beta_{samp}}\left[\log\left(1-D_\phi\left(s,a\right)\right)\right]
\end{split}
\end{equation}
Using this discriminator as a reward function, we can reward when data similar to the expert data is obtained as sample data, thus enabling efficient learning.

The algorithm based on the rewards and hyperparameters used in the experiment is shown in Algorithm \ref{DSQIL}. The hyperparameters were set to $\lambda_{demo}=1$, $\lambda_{samp}=1$, and $R=\frac{D\left(s,a\right)}{2}$ with the reward as half of the trained discriminator's output, based on the settings that performed best in SQIL, 1 reward for expert data and 0 reward for demo data. This setting allows us to limit the reward to between 0 and 1, close to the reward in SQIL. In this way, we can expect similar performance to SQIL in the early stages when learning is insufficient, and allow learning from sample data after some learning has progressed. Therefore, performance similar to or better than SQIL can be expected.

\section{Experimental Evaluation}

\subsection{Outline of experimentation}
To evaluate DSQIL, we compared DSQIL to SQIL using some empirical data: we evaluated Hopper-v2, Walker2d-v2, and HalfCheetah-v2 in three MuJoCo \cite{todorov2012mujoco} environments. Since the environment used in the experiment was a continuous-value control task, SAC \cite{haarnoja2018soft} was used as the agent in this case.

First, the policy is learned using SAC to obtain expert performance. The expert policy is used to generate a set of expert data to be stored in the replay buffer. The obtained expert policy performance is shown in Table \ref{expert_policy}. 

\begin{table}[t]
    \centering
    \caption{expert policy performance to provide expert data.}
    \label{expert_policy}
    \begin{tabular}{cc}
      \hline  
      Environments & Expert Performance \\ \hline
      Hopper-V3 & $3308.3\pm26.7$  \\
      Walker2D-V3 & $3897.7\pm31.6$ \\
      HalfCheetah-V3 & $5303.3\pm75.4$ \\ \hline
    \end{tabular}
\end{table}

\begin{table}[t]
  \centering
  \caption{SAC hyperparameters.}
  \label{SAC_hypara}
  \begin{tabular}{c c|c}
    & Parameters & Value \\ 
    \hline
    & Optimizer & Adam \\
    & Discount rate $\left(\gamma\right)$ & 0.99 \\
    Hyperparameters & $\alpha$ initial  & 0.2 \\
    & Actor learning rate  & $3e-4$ \\
    & Critic learning rate  & $3e-4$ \\
    & $\alpha$ learning rate  & $3e-4$ \\ 
    & Target update rate & $5e-3$ \\
    & Mini batch size & 64 \\
    \hline
    & Actor hidden dim & 256 \\
    & Actor hidden layers & 3 \\
    Architecture & Actor activation function & ReLU\\ 
    & Critic hidden dim & 256 \\
    & Critic hidden layers & 3 \\
    & Critic activation function & ReLU \\
  \end{tabular}
\end{table}

\begin{table}[t]
  \centering
  \caption{Discriminator hyperparameters.}
  \label{DIS_hypara}
  \begin{tabular}{c c|c}
    & Parameters & Value \\
    \hline
    & Optimizer & Adam \\
    & learning rate & $3e-4$ \\
    Hyperparameters & Mini batch size & 512 \\
    & Loss function & Binary Cross Entropy \\
    & replay buffer size($\beta_{samp}$) & 1e+6 \\
    & warm up & 1024 \\
    \hline
    & Hidden dim & 128 \\
    Architecture & Hidden layers & 3 \\
    & Activation function & Tanh \\
    & Last activation & Sigmoid \\
  \end{tabular}
\end{table}

\begin{table*}[t]
    \centering
    \caption{Performance (The maximum value for each trajectory is shown in bold.)}
    \label{experiment_data}
    \begin{tabular}{ccccc}
      \hline  
      Environments & trajectory & BC & SQIL & DSQIL(ours) \\ \hline
      Hopper-V3 & 2 & $2766.7\pm385.6$ & $\mathbf{3001.6\pm422.4}$ & $2886.9\pm741.3$ \\
      & 4 & $3010.8\pm460.3$ & $3113.9\pm365.6$ & $\mathbf{3202.3\pm187.1}$ \\
      & 8 & $3051.5\pm294.5$ & $3156.9\pm351.7$ & $\mathbf{3262.9\pm3.3}$ \\
      & 16 & $3164.7\pm269.8$ & $\mathbf{3274.9\pm6.1}$ & $3209.8\pm14.5$ \\
      & 32& $3241.1\pm141.0$ & $\mathbf{3287.6\pm1.9}$ & $3272.6\pm1.8$ \\ \hline
      Walker2D-V3 & 2 & $797.1\pm75.1$ & $2112.4\pm688.6$ & $\mathbf{3520.8\pm551.3}$ \\
      & 4 & $2001.0\pm1360.0$ & $3799.2\pm214.8$ & $\mathbf{3987.4\pm64.4}$ \\
      & 8 & $2110.0\pm1398.1$ & $\mathbf{3989.7\pm24.8}$ & $3929.9\pm15.5$ \\
      & 16 & $3571.7\pm588.6$ & $\mathbf{3911.9\pm18.3}$ & $3907.1\pm19.4$ \\
      & 32& $3756.5\pm236.2$ & $3905.7\pm11.1$ & $\mathbf{3921.9\pm41.9}$ \\ \hline
      HalfCheetah-V3 & 2 & $\mathbf{381.9\pm334.1}$ & $113.9\pm104.0$ & $99.0\pm103.1$\\ 
      & 4 & $599.3\pm306.9$ & $143.5\pm115.1$ & $\mathbf{1607.5\pm682.8}$ \\
      & 8 & $2622.4\pm342.0$ & $1874.4\pm2150.9$ & $\mathbf{4515.0\pm289.9}$ \\
      & 16 & $3228.4\pm671.2$ & $4361.4\pm86.8$ & $\mathbf{4499.4\pm105.2}$ \\
      & 32& $3350.5\pm491.4$ & $4636.8\pm161.9$ & $\mathbf{4713.5\pm69.3}$ \\ \hline
    \end{tabular}
\end{table*}

After the expert data is obtained, each method is trained. To study the effect of learning on the amount of expert data, the algorithms are trained on sets of $\left\{2, 4, 8, 16, 32\right\}$ as seen in \cite{papagiannis2022imitation}. In all environments, learning is performed in 500 steps, with one step acquiring one episode of sample data and storing it in the replay buffer. During training, sample and expert data are randomly extracted from the replay buffer at a ratio of 1:1 to be used for the discriminator in order to obtain a reward. At this time, the discriminator is trained simultaneously. All reported results correspond to performance measures obtained after testing the learner policy on 50 episodes. For agent we used SAC \cite{haarnoja2018soft}: a 3 layer MLP with ReLU activations and 256 hidden units. The discriminator is a 3 layer MLP with a sigmoid at the end in addition to tanh for activity and 128 hidden units. We trained all networks with the Adam optimizer \cite{kingma2014adam}.

We show other hyperparameters in Table \ref{SAC_hypara} and Table \ref{DIS_hypara}.

For each task, we compare the following three algorithms:

\begin{enumerate}
  \item BC
  \item SQIL based on SAC
  \item DSQIL based on SAC (ours)
\end{enumerate}

\subsection{Comparing the scores}
Evaluation results are shown in Table \ref{experiment_data}. In each environment, DSQIL outperforms BC. For relatively simple tasks such as the Hopper and Walker2d environments, SQIL and DSQIL have comparable performance if sufficient expert data is available. On the other hand, in complex environments with more information required, such as the HalfCheetah environment, DSQIL performs better than SQIL regardless of the amount of expert data. The performance difference is especially large when there is less expert data. However, for simple tasks such as the Hopper environment, there is no performance difference depending on the amount of expert data, and DSQIL shows lower performance than SQIL with a small amount of expert data.

\subsection{Comparing the speed of learning} \label{com_score}
\begin{figure*}[h]
    \centering
    
    \begin{minipage}[h]{0.6\linewidth}
        \centering
        
        \includegraphics[keepaspectratio, scale=0.45]{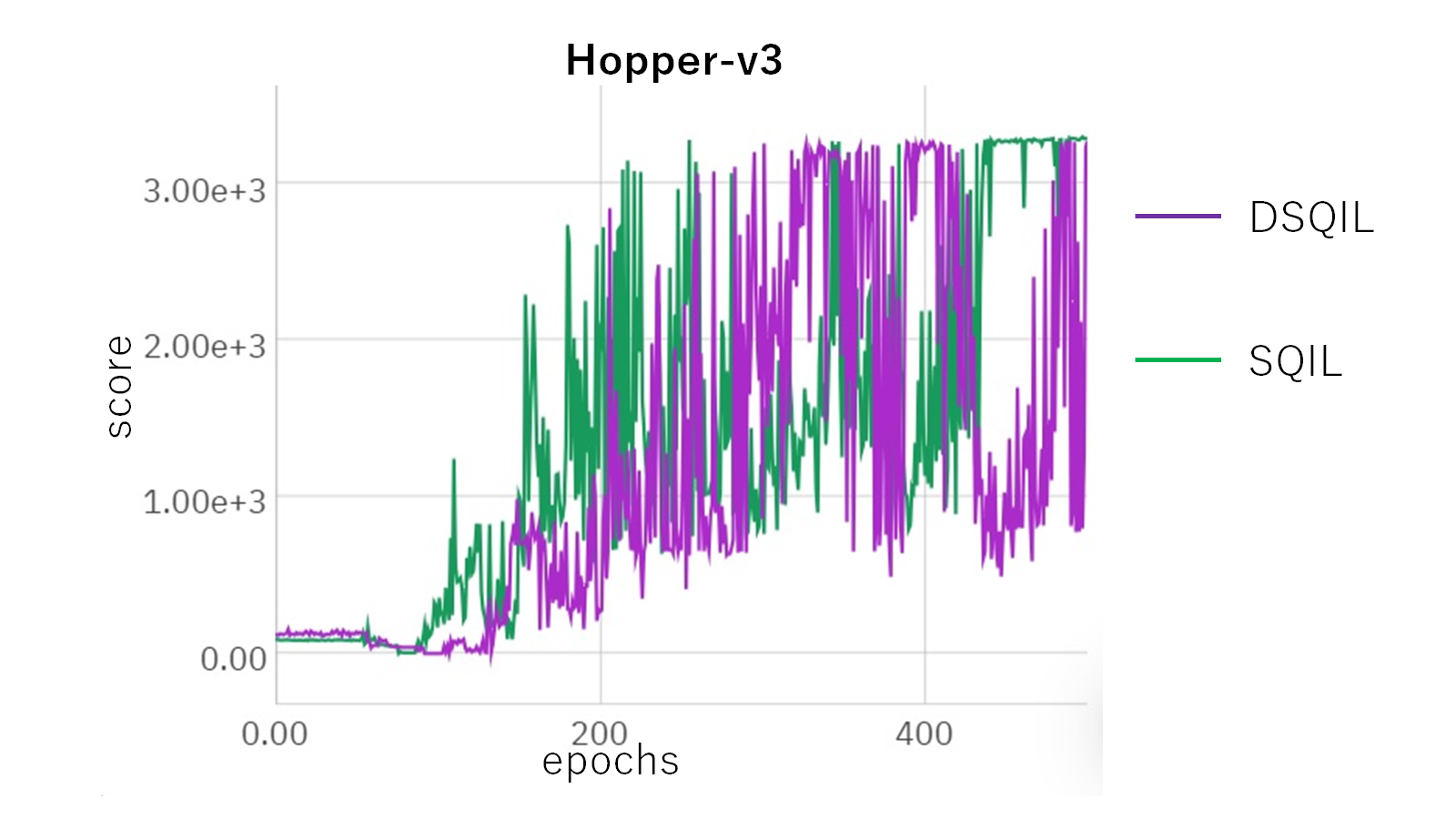}
        \subcaption{Hopper-v3}
    \end{minipage}
    
    \begin{minipage}[h]{0.6\linewidth}
        \centering
        
        \includegraphics[keepaspectratio, scale=0.45]{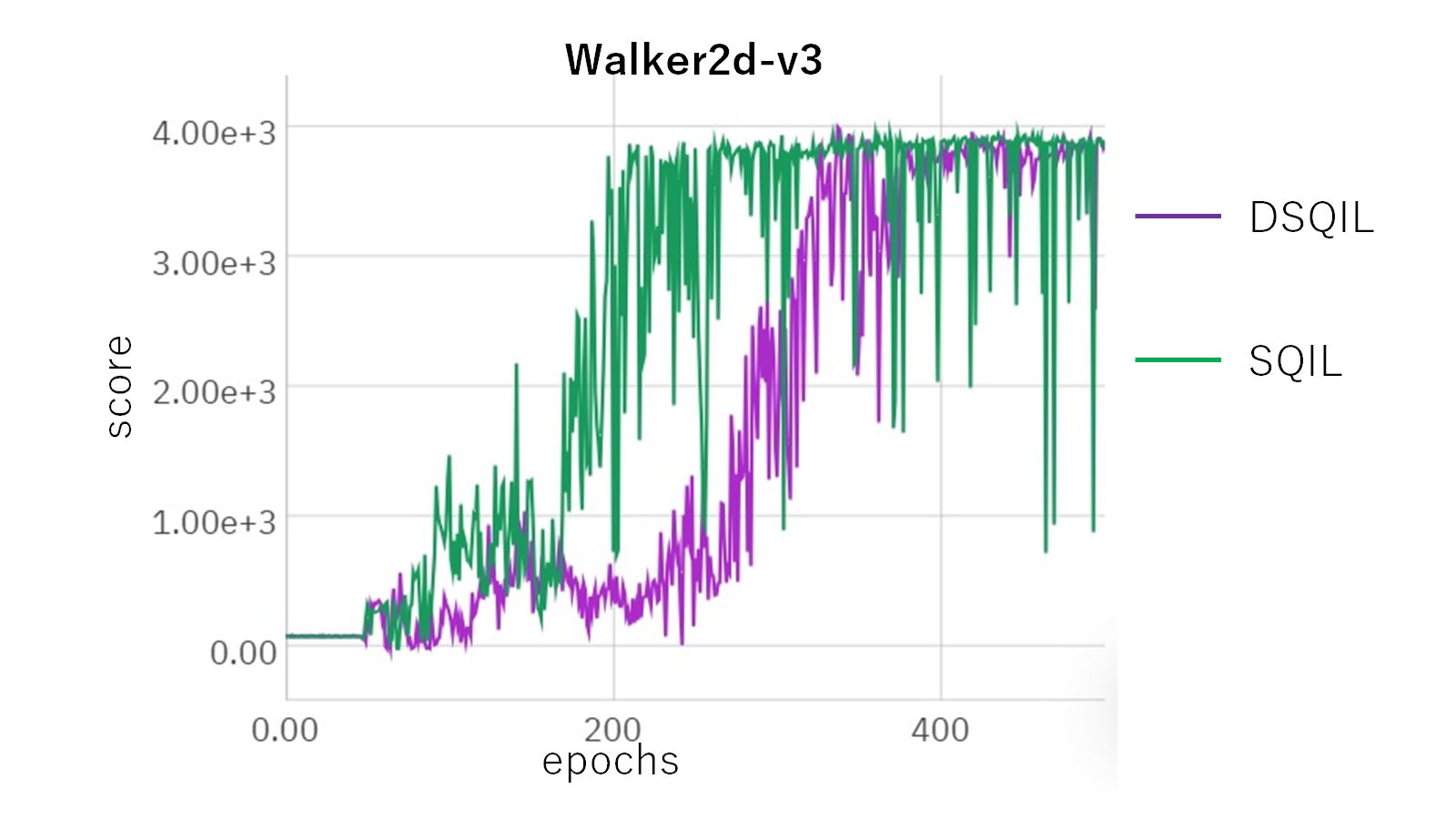}
        \subcaption{Walker2d-v3}
    \end{minipage}

    \begin{minipage}[h]{0.6\linewidth}
        \centering
        
        \includegraphics[keepaspectratio, scale=0.45]{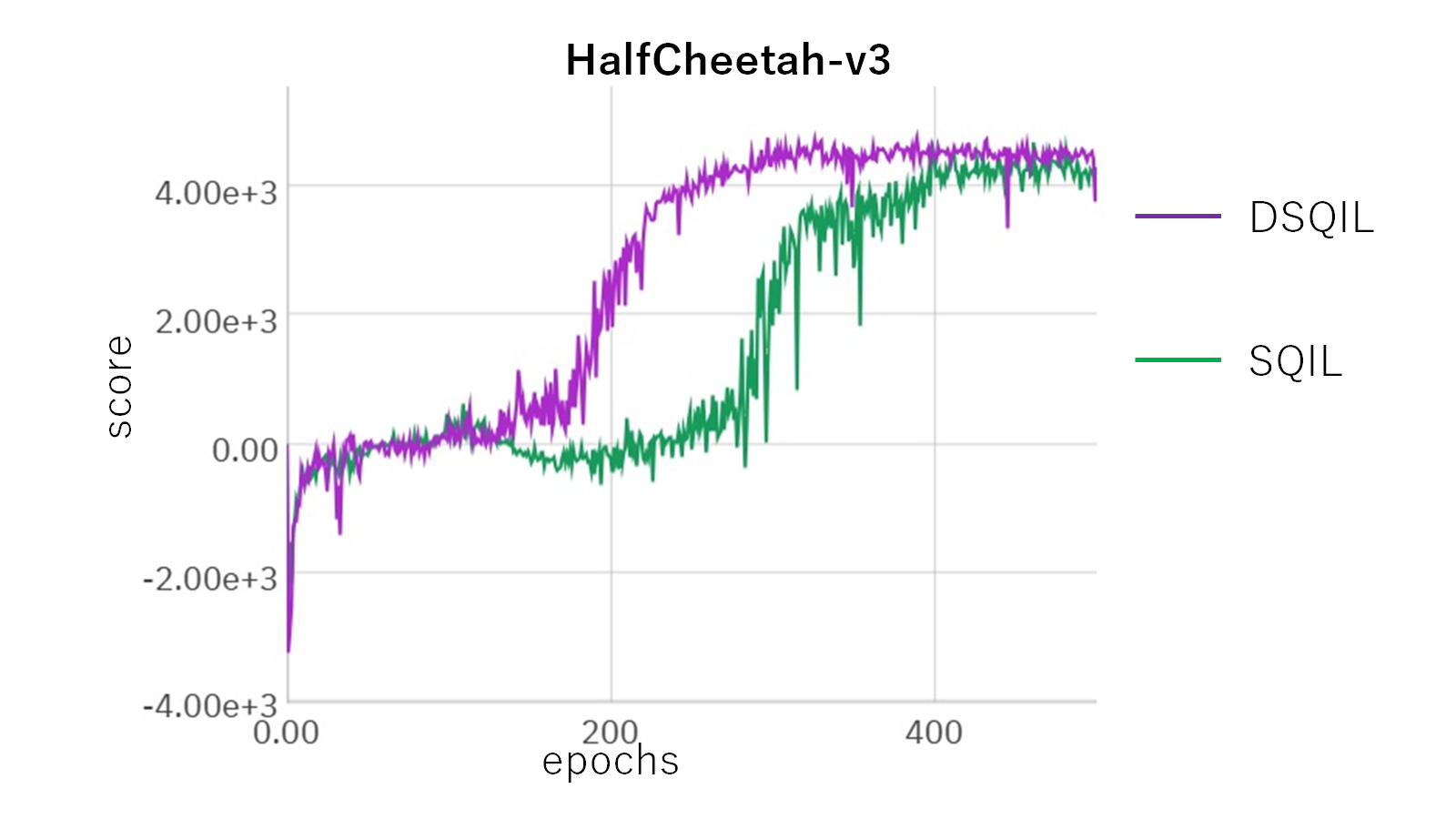}
        \subcaption{HalfCheetah-v3}
    \end{minipage}
    \caption{Comparison of the average score per epoch for each environment for 32 episodes of expert data.}
    \label{score}
\end{figure*}

The training for DSQIL and SQIL is shown in Figure \ref{score}. It shows the learning for each environment given 32 episodes of expert data. At the end of an epoch, 5 episodes are tested using the network learned up to that point, and the average of the results is obtained. Figure \ref{score} shows the evolution of these values obtained with SQIL and DSQIL. It can be seen that in the relatively simple Hopper and Walker2d environments, the SQIL learns faster than the DSQIL. This is due to the fact that training a discriminator requires a certain number of steps. If the discriminator is not well trained, it is possible that the learning could be affected by giving unjustified rewards to the sample data. On the other hand, in the complex HalfCheetah environment, DSQIL learns faster than SQIL. This indicates that in complex environments, the effect of being able to learn from sample data is greater than the loss incurred during the training period of the discriminator.

\subsection{Comparing the rewards}
\begin{figure*}[h]
    \centering
    \begin{minipage}[h]{0.45\linewidth}
        \centering
        
        \includegraphics[keepaspectratio, scale=0.3]{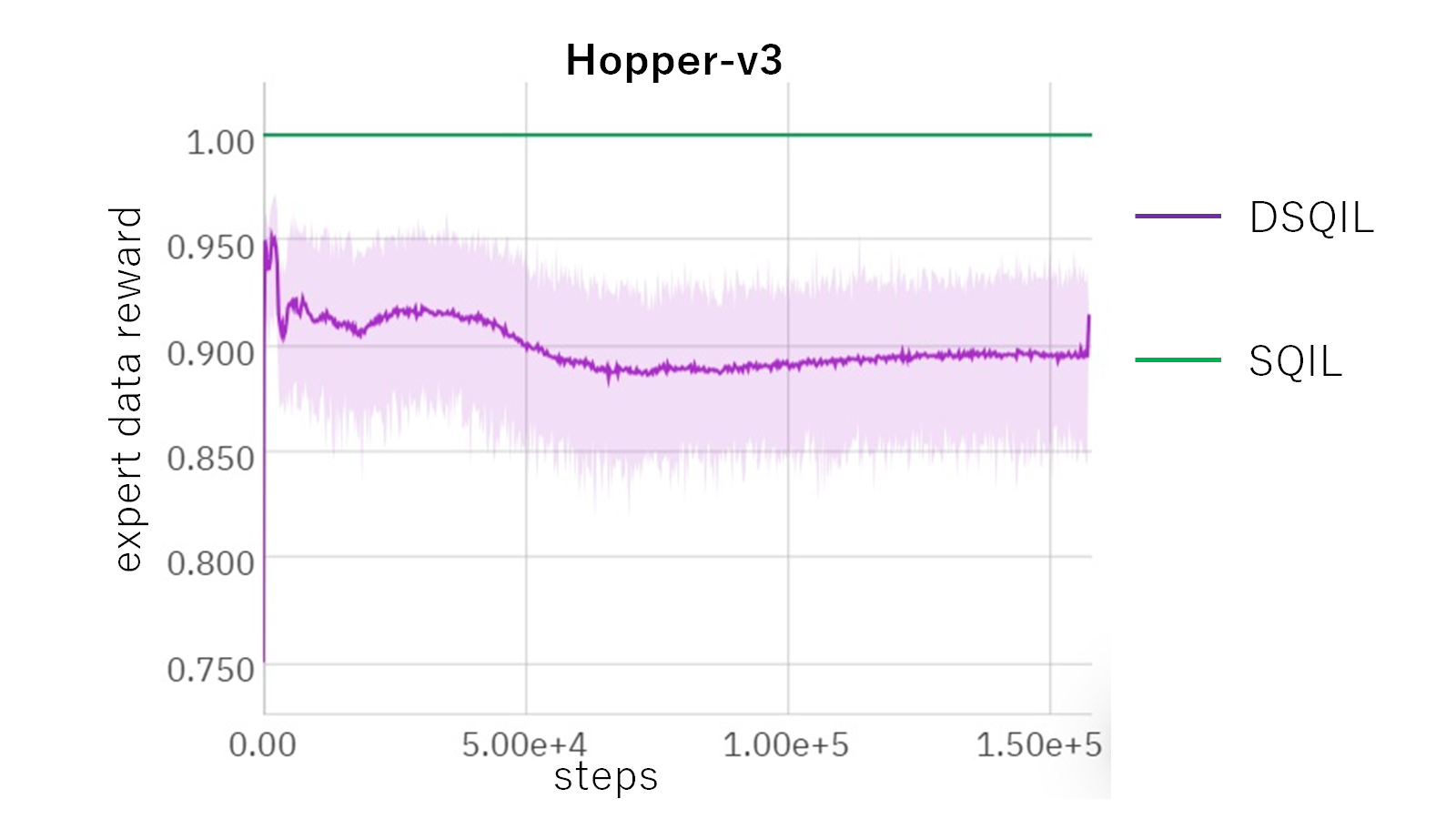}
        \subcaption{Expert data rewards in Hopper-v3}

    \end{minipage}
    \begin{minipage}[h]{0.45\linewidth}
        \centering
        
        \includegraphics[keepaspectratio, scale=0.3]{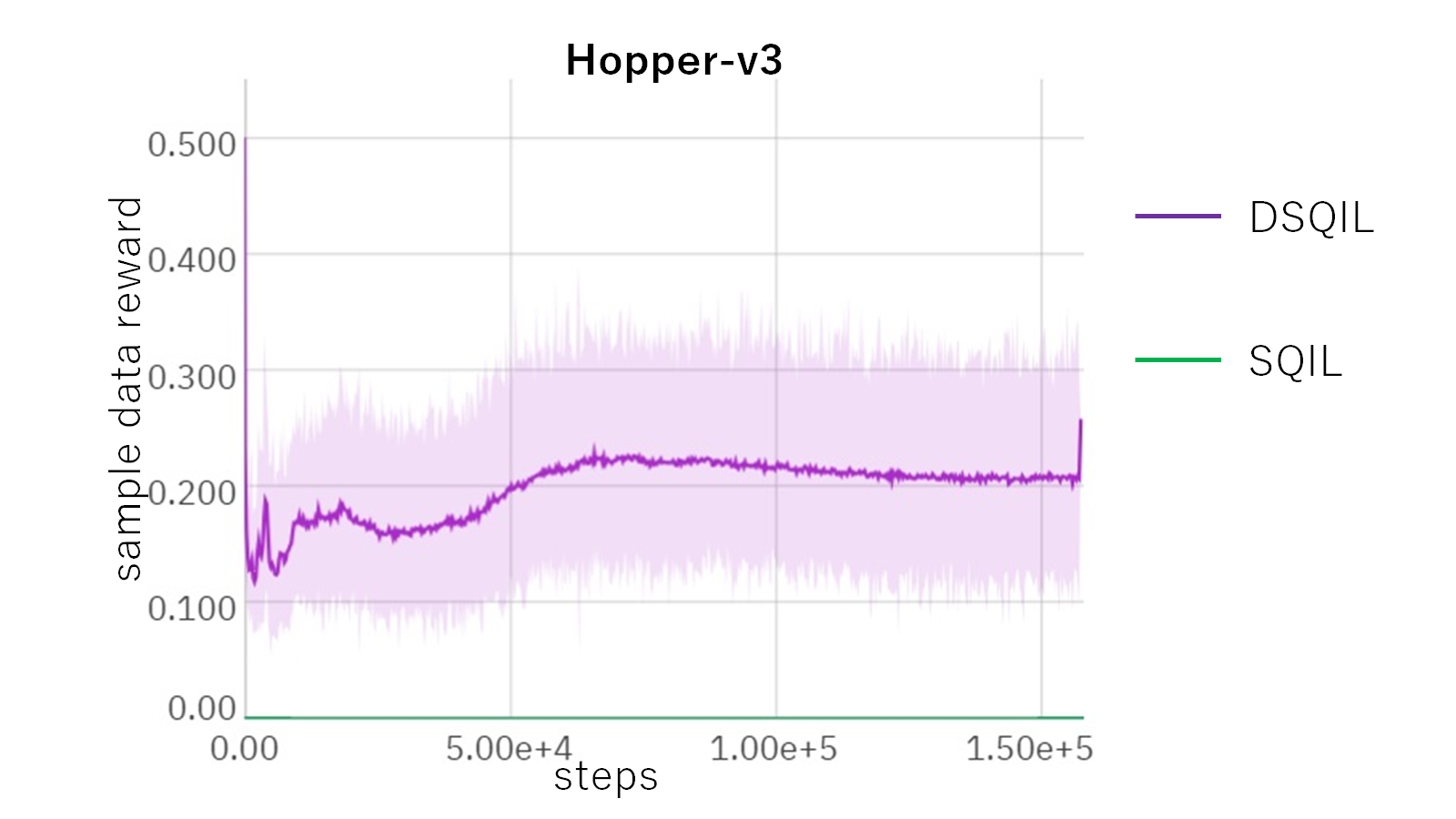}
        \subcaption{Sample data rewards in Hopper-v3}
    \end{minipage}
    \begin{minipage}[h]{0.45\linewidth}
        \centering
        
        \includegraphics[keepaspectratio, scale=0.3]{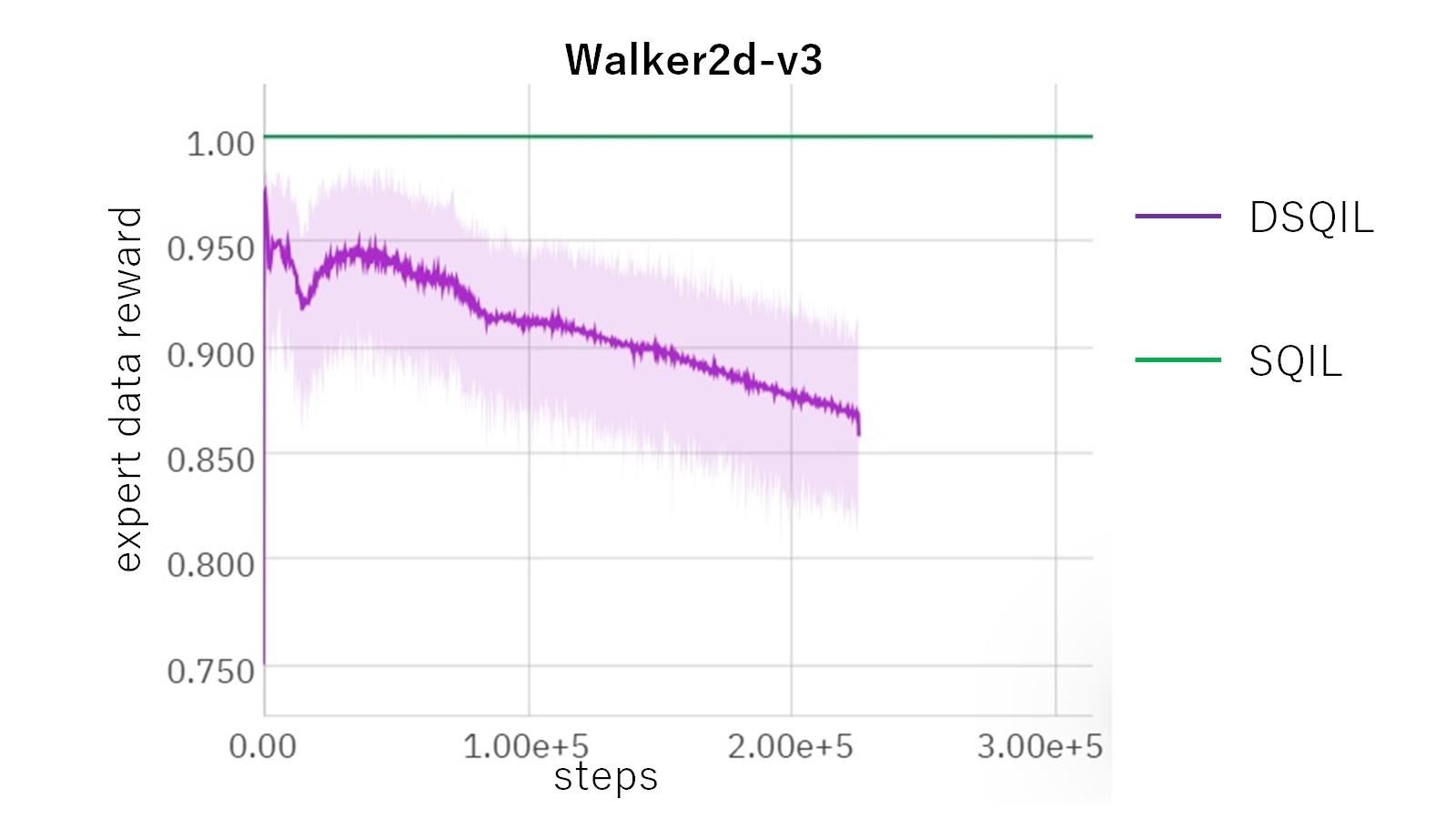}
        \subcaption{Expert data rewards in Walker2d-v3}

    \end{minipage}
    \begin{minipage}[h]{0.45\linewidth}
        \centering
        
        \includegraphics[keepaspectratio, scale=0.3]{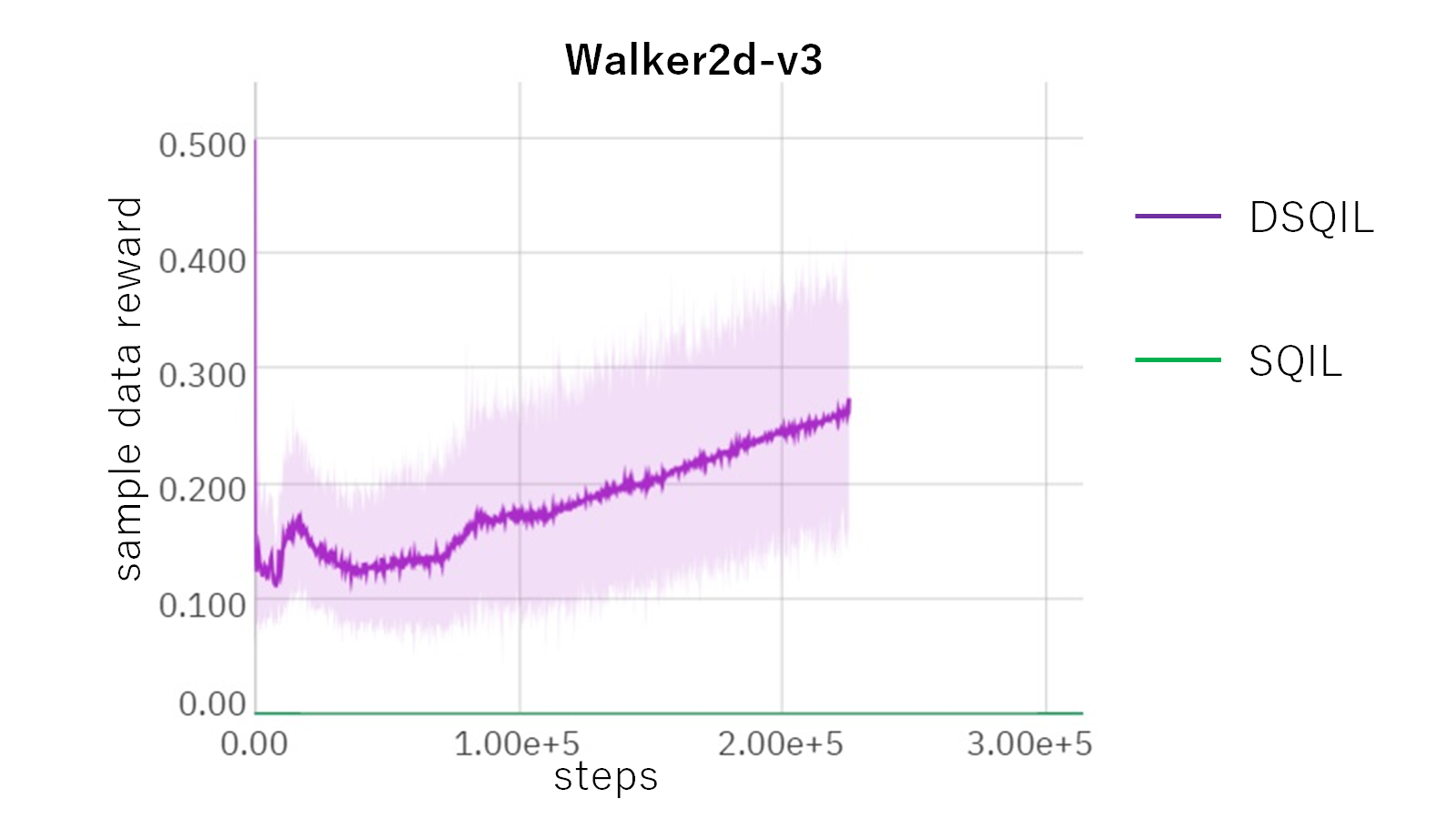}
        \subcaption{Sample data rewards in Walker2d-v3}
    \end{minipage}
    \begin{minipage}[h]{0.45\linewidth}
        \centering
        
        \includegraphics[keepaspectratio, scale=0.3]{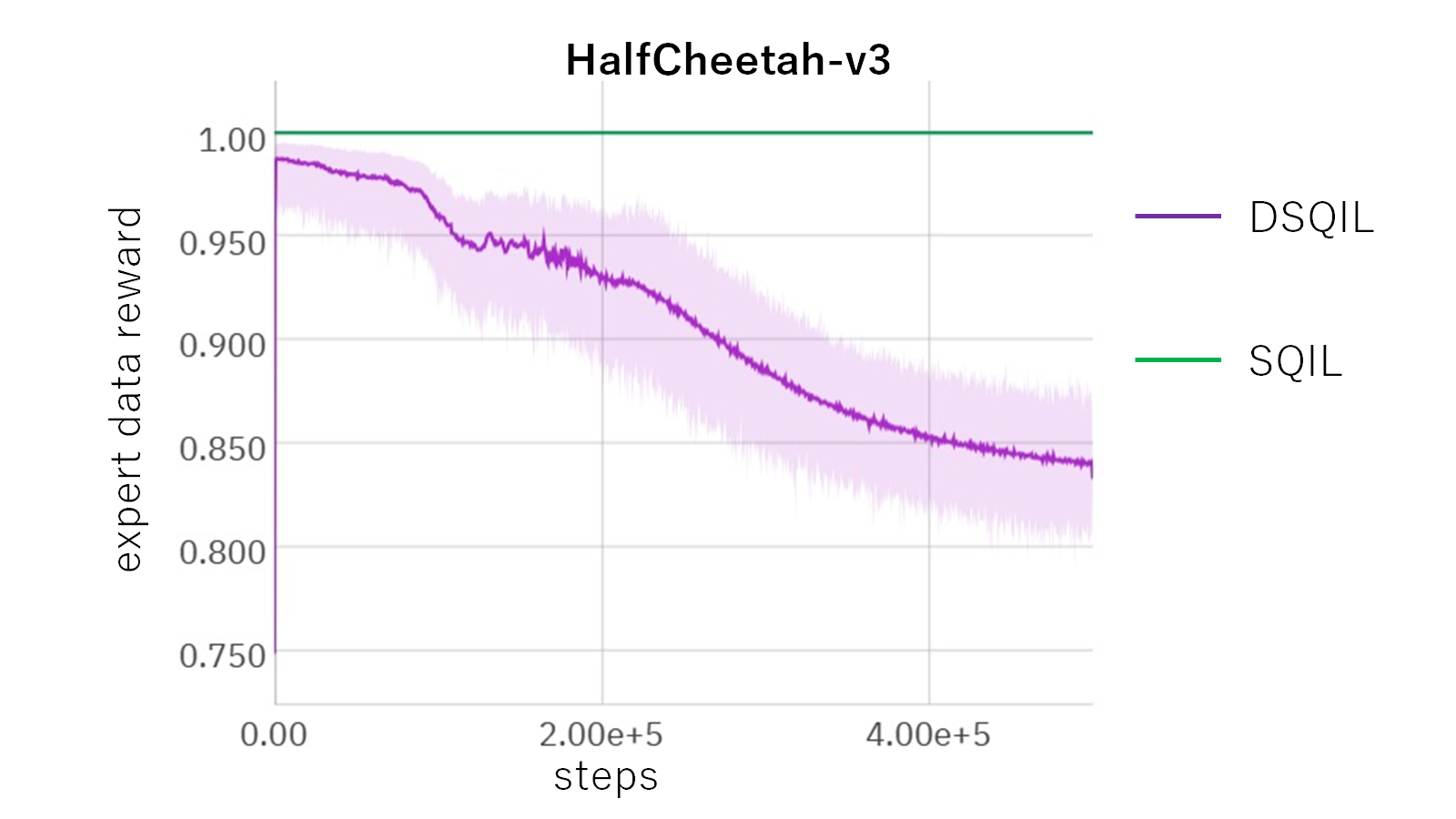}
        \subcaption{Expert data rewards in HalfCheetah-v3}

    \end{minipage}
    \begin{minipage}[h]{0.45\linewidth}
        \centering
        
        \includegraphics[keepaspectratio, scale=0.3]{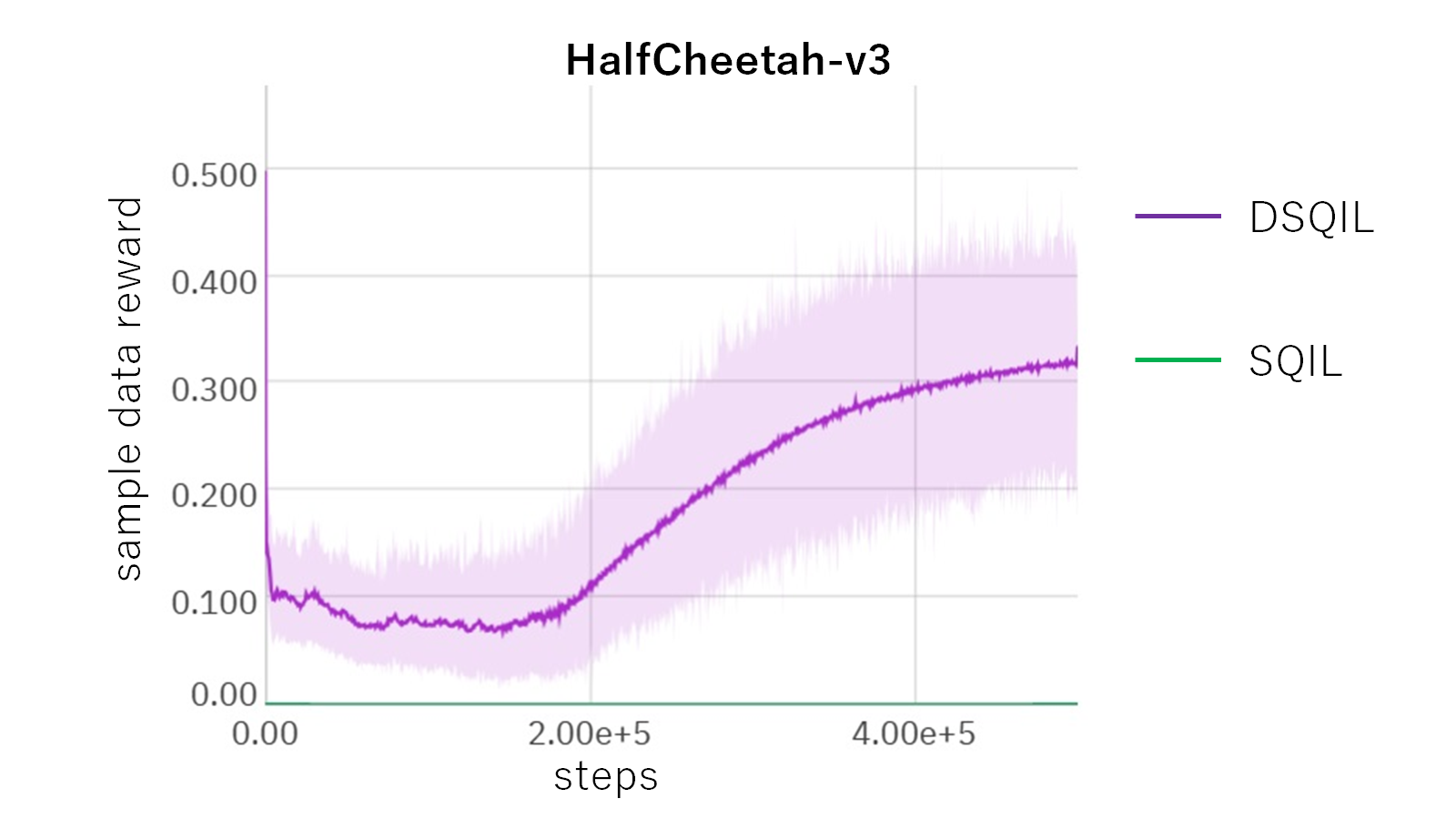}
        \subcaption{Sample data rewards in HalfCheetah-v3}
    \end{minipage}
    \caption{Comparison of expert data and sample data rewards for each environment with 32 episodes of expert data.}
    \label{rewards}
\end{figure*}

A comparison between DSQIL and SQIL for each reward transition for the sample and expert data in each environment is shown in Figure \ref{rewards}. This shows the evolution of rewards during step-by-step learning when 32 episodes of expert data are given as in Chapter \ref{com_score}. As we designed, SQIL continues to give fixed values, while DSQIL shows variable rewards. In particular, in the HalfCheetah environment, it is noticeable that the reward for the sample data increases as the learning progresses. It can be seen that the rewards are given to data that is close to the expert data found in the sample data and used for learning. On the other hand, we observed a decrease in the reward for expert data. Although no effect of this was observed in the experiment, it is possible that the accuracy of the data decreases when the number of learning epochs is increased, or that the data behaves differently from the expert data.

\section{Conclusion}

In this paper, we propose Discriminator Soft Q Imitation Learning (DSQIL) as a data-efficient imitation learning method. In contrast to the conventional method SQIL, we show that DSQIL can provide more detailed rewards for state-action pairs by using a reward function instead of a constant reward. The method incorporates the idea of a GAN discriminator in the reward function and was evaluated in three experiments with MuJoCo.

The experiments confirm that DSQIL performs as well as or better than conventional imitation learning. Especially in complex environments, DSQIL outperforms SQIL in both data efficiency and learning efficiency. On the other hand, in certain environments, the rewards for both expert and sample data tended to converge to similar values as learning progressed. This can be an advantage when seeking higher performance from the expert data, but a disadvantage when seeking performance comparable to the expert data.

Based on the above considerations, it is necessary to examine the extent to which discriminator accuracy affects learning. It is also worth continuing research in terms of verifying performance in more complex environments and designing reward functions that improve performance in simple environments.

\bibliographystyle{ieicetr}
\bibliography{ref.bib}


\profile[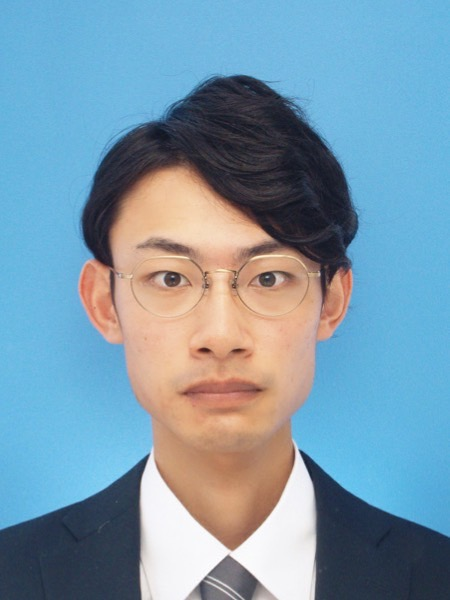]{Ryoma Furuyama}{He received a B.S. degree from Kanazawa University in 2022. He is now a M.S. student studying reinforcement learning.}

\profile[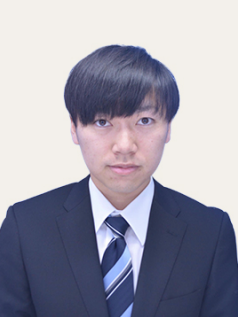]{Daiki Kuyoshi}{He received a M.S. degree from Kanazawa University in 2022. He is interested in reinforcement learning}

\profile[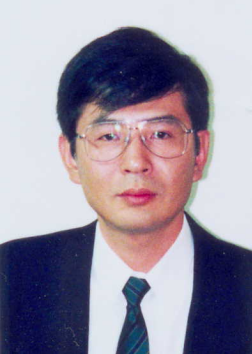]{Satoshi Yamane}{He received B.S.,M.S and Ph.D. degrees from Kyoto University. Now he is a professor of Kanazawa University. He is interested in formal verification of real-time and distributed computing.}

\end{document}